\documentclass[conference]{IEEEtran}
\IEEEoverridecommandlockouts

\usepackage{cite}
\usepackage{amsmath,amssymb,amsfonts}
\usepackage{algorithm}
\usepackage{algpseudocode}
\usepackage{graphicx}
\usepackage{booktabs}
\usepackage{multirow}
\usepackage{xcolor}
\usepackage{url}
\usepackage[hidelinks]{hyperref}
\usepackage{tikz}
\usepackage{tabularx}
\usepackage{xfrac}

\algrenewcommand\alglinenumber[1]{\scriptsize #1:}
\usetikzlibrary{arrows.meta,positioning,shapes.geometric,fit,backgrounds,calc}

\newcommand{\HFW}{\textsc{HFW}}
\newcommand{\Mmat}{\mathbf{M}}
\newcommand{\etasym}{\eta}

\begin{document}


\title{Where to Bind Matters: Hebbian Fast Weights in Vision Transformers for Few-Shot Character Recognition\\}

\author{
\IEEEauthorblockN{
Gavin Money, Sindhuja Penchala, Jiacheng Li, Noorbakhsh Amiri Golilarz
}
\IEEEauthorblockA{
Department of Computer Science, The University of Alabama, Tuscaloosa, AL, USA\\
\{gcmoney, spenchala\}@crimson.ua.edu, jiachengli@ieee.org, noor.amiri@ua.edu
}
}






\maketitle

\begin{abstract}
Standard transformer architectures learn fixed slow-weight representations during training and lack mechanisms for rapid adaptation within an episode. In contrast, Biological neural systems address this through fast synaptic updates that form transient associative memories during inference, a property known as Hebbian plasticity. In this paper, we conduct an empirical study of Hebbian Fast-Weight (HFW) modules integrated into multiple transformer backbones, including ViT-Small, DeiT-Small, and Swin-Tiny. We evaluate six model variants  ViT, DeiT, Swin, ViT-Hebbian, DeiT-Hebbian, and Swin-Hebbian on 5-way 1-shot and 5-way 5-shot classification tasks using the Omniglot benchmark under a Prototypical Network meta learning framework. We propose a single module placement strategy for Swin-Tiny in which one HFW module is applied to the final stage feature map after all hierarchical stages have completed. This design avoids the training instability caused by placing separate Hebbian modules at each stage and achieves the highest test accuracy across all six models ($96.2\%$ at 1-shot; $99.2\%$ at 5-shot), outperforming its non-Hebbian baseline by $+0.3$ percentage points at 1-shot. We analyze the interaction between Swin's shifted window inductive bias and episode level Hebbian binding, discuss why per-block placement fails for ViT and DeiT variants in a low data regime, and situate the results within the wider literature on fast/slow weight meta learning.


\end{abstract}

\begin{IEEEkeywords}
Hebbian learning, fast weights, few-shot learning, Swin Transformer,
Omniglot, meta-learning, Prototypical Networks, synaptic plasticity
\end{IEEEkeywords}

\section{Introduction}
\label{sec:intro}

The ability to recognize a novel concept from a single or handful of examples is a hallmark of human cognition, yet remains a fundamental challenge for deep neural networks. Standard vision transformers \cite{dosovitskiy2020vit} \cite{touvron2021deit} excel in large scale supervised learning but rely entirely on gradient descent over massive datasets to encode knowledge into fixed slow weights, parameters that are frozen during inference and therefore incapable of adapting to new information seen within a single episode. Despite these advances, these models rely heavily on static parameterization through gradient based optimization, where knowledge is encoded into fixed slow weights that remain unchanged during inference. This design inherently limits their ability to adapt to new information in real time, particularly in few-shot or one-shot learning scenarios.

Biological learning operates across two timescales simultaneously \cite{hebb1949}. Slow synaptic consolidation corresponds roughly to offline gradient based training, but fast Hebbian style plasticity allows the brain to bind novel associations from a single exposure without overwriting previously learned knowledge \cite{ba2016fastweights}. This dual timescale view motivates fast-weight memory: a transient, episode level memory that is written from key-value co activations and erased at the start of each new episode \cite{schlag2021linear} \cite{munkhdalai2018hebbian}. This dual timescale learning paradigm provides a compelling framework for addressing the adaptability limitations of modern neural architectures. Specifically, fast-weight memory mechanisms ephemeral and context dependent allow models to temporarily encode associations within an episode without interfering with previously learned knowledge. Such mechanisms have been explored in prior works on fast weights and meta learning \cite{ba2016fastweights} \cite{schlag2021linear} \cite{munkhdalai2018hebbian}, demonstrating their potential for rapid task adaptation.

Recent theoretical work \cite{schlag2021linear} has proven that linear attention in standard transformers is mathematically equivalent to executing an outer-product Hebbian update, revealing that transformers already implement a form of implicit fast-weight programming. This insight motivates making the Hebbian update explicit, learnable, and architecturally distinct from the slow-weight attention pathway, enabling the network to learn not just what associations to form but how quickly to form and forget them.

\vspace{0.1cm}

More broadly, this perspective aligns with the emerging direction of neurocognitive inspired intelligence \cite{golilarz2025neuro} \cite{golilarz2025cognitive}, which advocates for embedding biologically grounded mechanisms, such as fast associative memory into deep learning systems to achieve adaptive and context aware intelligence beyond what can be attained through scaling alone. The primary objective of this study is to investigate the effectiveness of the HFW memory mechanisms in enhancing the adaptability of transformer based architectures for few-shot learning. Specifically, this work aims to:

\begin{itemize}
  \item Design a Hebbian Fast-Weight module with learnable plasticity rate $\etasym$, temporal decay $\lambda$, gated sigmoid output, and Frobenius norm stabilization for stable memory updates.
  
  \item Integrate the proposed \HFW\ module into multiple transformer backbones, including ViT-Small (per-block), DeiT-Small (per-block), and Swin-Tiny (single final stage module), resulting in six model variants.
  
  \item Evaluate the performance of these models on Omniglot 5-way 1-shot and 5-shot classification tasks using Prototypical Networks as the metric based learning framework.
  
  \item Analyze the impact of architectural placement of the HFW module and demonstrate that a single final stage integration in Swin-Tiny provides improved stability and performance compared to per-block variants.
\end{itemize}



The remainder of this paper is organized as follows. Section \ref{sec:related} reviews related work on few-shot learning, transformer architectures, and fast-weight memory mechanisms. Section \ref{sec:arch} presents the proposed architecture, including the Hebbian Fast-Weight module and its integration into different backbones. Section \ref{sec:setup} describes the experimental setup, datasets, and evaluation protocol. Section \ref{sec:results} reports the results and quantitative analysis. Section \ref{sec:analysis} provides a detailed discussion on why the Swin-Hebbian architecture outperforms per-block Hebbian variants. Section \ref{sec:discussion} discusses broader implications, limitations, and future directions. Finally, Section \ref{sec:conclusion} concludes the paper.

\section{Related Work}
\label{sec:related}

The problem of few-shot learning has been extensively studied across meta-learning, memory-augmented models, and transformer-based architectures. In this section, we summarize key developments in fast-weight mechanisms, few-shot learning paradigms, and biologically plausible plasticity. These works form the basis for our proposed approach.

\subsection{Fast Weights and Hebbian Memory}
The concept of complementary fast and slow weights traces to Hinton and Plaut \cite{hintonplaut1987} and Schmidhuber's fast-weight controllers \cite{schmidhuber1992}. The modern revival by Ba et al. \cite{ba2016fastweights} demonstrated that fast weights attending to recent activations improve sequential recall. Munkhdalai and Trischler \cite{munkhdalai2018hebbian} extended this explicitly to meta learning, showing that a Hebbian fast-weight term coupled to a slow-weight backbone improves 5-way accuracy on Omniglot, the most direct antecedent to this paper. Schlag et al. \cite{schlag2021linear} then unified the fast-weight programmer formalism with linear attention, proving that the outer product write rule is the Hebbian rule at the heart of modern transformers. Irie et al. \cite{irie2021recurrent} extended this to recurrent fast-weight programmers, and Chaudhary \cite{chaudhary2025robust} recently demonstrated that Hebbian and gradient based plasticity together yield robust in-context memory in transformers.

\subsection{Few-Shot Learning}
The meta learning literature offers several complementary approaches to few-shot recognition. Metric based methods, such as Matching Networks \cite{vinyals2016matching}, Prototypical Networks \cite{snell2017proto}, and Relation Networks \cite{sung2018relation}, embed support and query images into a shared space and classify by proximity to class prototypes or learned similarity. Optimization based methods such as MAML \cite{finn2017maml} and Reptile \cite{nichol2018reptile} learn initializations that can be rapidly fine tuned. Memory augmented approaches including MANN \cite{santoro2016mann} and Meta Networks \cite{munkhdalai2017meta} maintain explicit external memories. The Hebbian fast-weight approach bridges metric based and memory augmented paradigms: the backbone's slow weights define an episode invariant embedding space while the fast-weight matrix dynamically binds support set identity associations during inference.

\subsection{Transformers in Few-Shot Vision}
SNAIL \cite{mishra2018snail} first demonstrated that temporal convolutional attention could serve as a general meta learner. Ye et al. \cite{ye2020feat} used set to set transformer adaptation of embeddings, while CrossTransformers \cite{doersch2020cross} used spatially aware cross attention between support and query. Swin Transformer \cite{liu2021swin} introduced a hierarchical, shifted window attention design that naturally captures multi scale structure while maintaining linear complexity in sequence length, properties that are well suited to character recognition benchmarks like Omniglot.

\subsection{Biologically Plausible Plasticity}
Differentiable plasticity \cite{miconi2018plasticity} trains Hebbian coefficients end to end. Meta learning through Hebbian plasticity in random networks \cite{najarro2020hebbian} showed that randomly initialized networks augmented with learnable Hebbian plasticity rules can match trained baselines. Shervani-Tabar and Rosenbaum \cite{shervani2023} demonstrated meta learning of biologically plausible plasticity rules using random feedback pathways. These works collectively establish that explicitly incorporating Hebbian dynamics into deep networks is both feasible and beneficial.

\section{Architecture}
\label{sec:arch}
In this section, we first present the Hebbian Fast Weight module, including its memory update and retrieval mechanism, and then describe its integration into ViT, DeiT, and Swin architectures. The section concludes by comparing these integration strategies to clarify the architectural basis for their different empirical behaviors.

\subsection{Hebbian Fast Weight Module (\HFW)}
\label{sec:hfw}
\begin{figure}[t]
    \centering
    \includegraphics[width=\columnwidth]{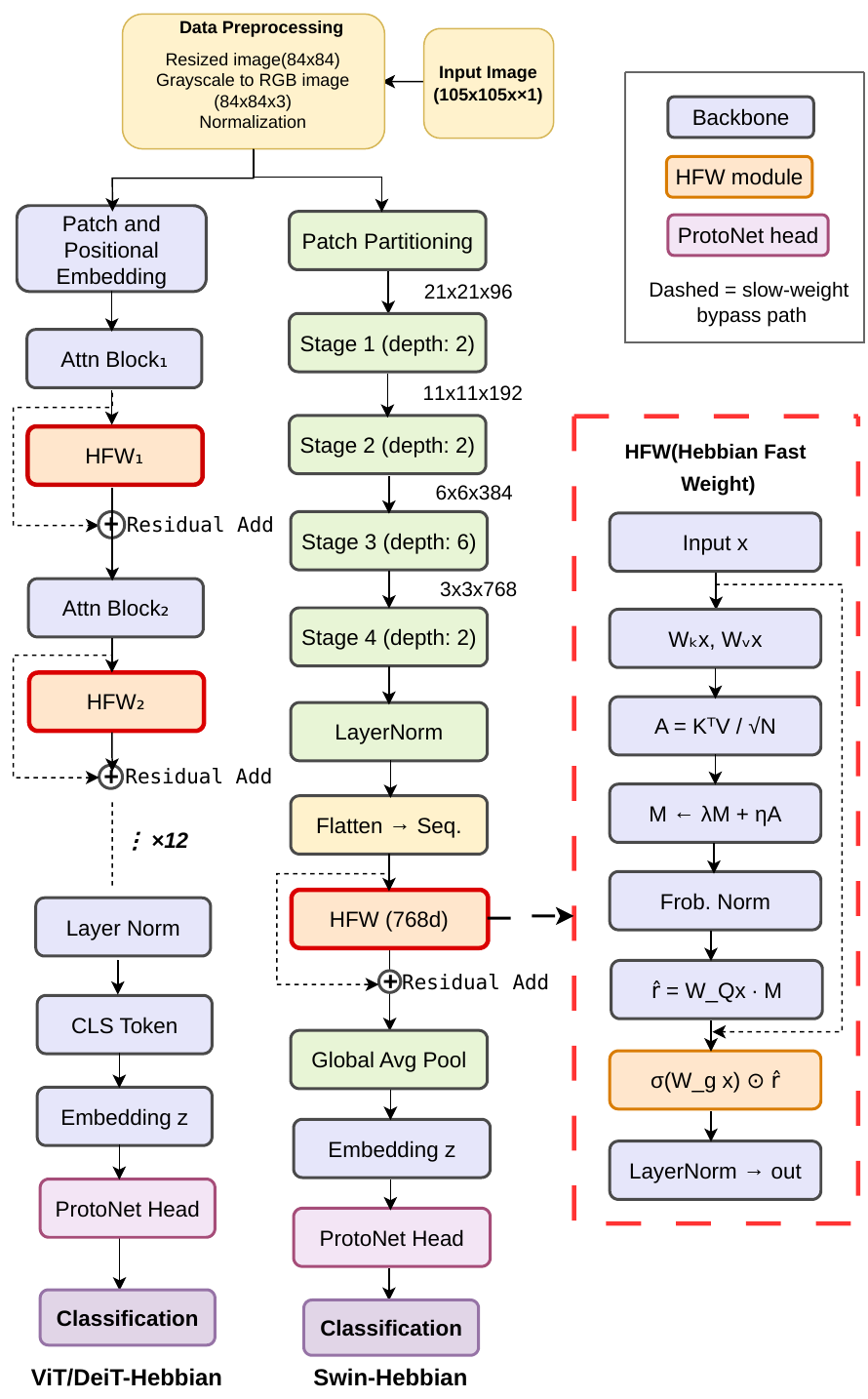}
    \caption{Proposed Hebbian-enhanced Transformer architectures. \textbf{Left:} ViT/DeiT-Hebbian inserts an HFW module in parallel with the attention blocks (repeated $\times 12$), with residual bypass connections. \textbf{Center:} Swin-Hebbian applies a single HFW module after the hierarchical stages, LayerNorm, and sequence flattening, followed by pooling and ProtoNet classification. \textbf{Right:} Internal HFW computation showing associative memory update, retrieval, gating, and normalization; dashed paths indicate bypass connections and input-dependent gating.}
    \label{fig:architecture}
\end{figure}

The overall architecture of the proposed framework is illustrated in Fig. \ref{fig:architecture}. The Hebbian Fast-Weight module is designed to augment the backbone with dynamic memory updates and retrieval.

The Hebbian Fast Weight module introduces a transient associative memory mechanism to enhance the adaptability of transformer based architectures in few shot learning. In contrast to the conventional slow weight pathway, which stores knowledge in fixed parameters learned during training, the proposed module dynamically forms episode specific associations during inference. This design allows the model to rapidly bind newly observed support information without modifying the backbone weights, thereby providing an additional fast memory pathway for within episode adaptation.

The HFW module maintains a transient associative memory matrix
\begin{equation}
M \in \mathbb{R}^{B \times H \times d_h \times d_h},
\label{eq:hfw_memory}
\end{equation}
where $B$ denotes the batch size, $H$ is the number of attention heads, and $d_h = d/H$ is the dimension of each head. The memory matrix $M$ is initialized to zero at the beginning of each forward pass and is not treated as a trainable parameter. Instead, it is constructed online from the current input sequence and serves as an episode level memory for temporary associative binding.

Given an input token sequence $x \in \mathbb{R}^{B \times N \times d}$, the module first computes key, value, and query representations through three learned linear projections:
\begin{equation}
K = W_K x,\quad V = W_V x,\quad Q = W_Q x
\label{eq:hfw_qkv}
\end{equation}
where $W_K$, $W_V$, and $W_Q$ denote the projection matrices for keys, values, and queries, respectively.

To construct the associative memory, the module computes a co-activation matrix from the key and value representations:
\begin{equation}
A = \mathrm{clamp}\!\left(\frac{K^\top V}{\sqrt{N}}, -\delta, +\delta\right)
\label{eq:hfw_assoc}
\end{equation}
where $\delta$ is a clipping threshold that prevents excessively large activations and improves training stability. Based on this associative signal, the memory matrix is updated according to
\begin{equation}
M = \lambda M + \eta A,\qquad
M \leftarrow \frac{M}{\|M\|_F + \epsilon}
\label{eq:hfw_update}
\end{equation}

where $\eta$ is the plasticity rate, $\lambda$ is the temporal decay factor, $\|\cdot\|_F$ denotes the Frobenius norm, and $\epsilon$ is a small constant for numerical stability. The first term preserves previously accumulated associations with controlled decay, while the second term incorporates newly observed associations from the current episode. The normalization step further stabilizes the magnitude of the memory matrix during training.

The plasticity rate and decay factor are both learnable scalar parameters constrained to the interval $(0,1)$ through sigmoid activations:
\begin{equation}
\eta = \sigma(\eta_{\mathrm{logit}})\cdot \eta_{\max},\qquad
\lambda = \sigma(\lambda_{\mathrm{logit}})
\label{eq:hfw_eta_lambda}
\end{equation}
where $\eta_{\max}$ sets the maximum allowable plasticity strength. This parameterization allows the model to learn how strongly new associations should be written into memory and how long those associations should persist within an episode.

After the memory matrix has been updated, the module retrieves associative information by applying the query representation to the learned memory:
\begin{equation}
\hat{r} = Q M
\label{eq:hfw_retrieval}
\end{equation}
where $\hat{r}$ denotes the retrieved fast memory response. To regulate the influence of this retrieved signal, the module applies an element wise gating mechanism:
\begin{equation}
\mathrm{out} = \mathrm{LayerNorm}\!\left(\sigma(W_g x) \odot \hat{r}\right)
\label{eq:hfw_output}
\end{equation}
where $W_g$ is a learned projection matrix, $\sigma(\cdot)$ denotes the sigmoid activation function, and $\odot$ represents element wise multiplication. This gating operation allows the network to adaptively control how much of the retrieved associative memory should contribute to the final output, thereby balancing the fast memory pathway against the standard slow weight representation.

It is important to note that the memory matrix $M$ itself is transient and is therefore not stored in the model checkpoint. The checkpoint only stores the learned parameters that govern memory formation and retrieval, including $W_K$, $W_V$, $W_Q$, $W_g$, and the logits associated with $\eta$ and $\lambda$. Consequently, the model preserves the strategy for constructing fast associations rather than the associations formed in any particular episode.


\subsection{Per-Block Hebbian Placement in ViT and DeiT}
As shown in Fig. \ref{fig:architecture}, the Hebbian Fast-Weight module is integrated into different transformer architectures at specific stages, interacting with intermediate feature representations during forward propagation.

For the ViT-Small and DeiT-Small backbones, the HFW module is inserted into every transformer block, resulting in a per-block fast memory design. Let $x_\ell$ denote the input to the $\ell$th transformer block. The output of the block is augmented by an additional Hebbian pathway, yielding
\begin{equation}
x_{\ell+1} = \mathrm{TransformerBlock}_\ell(x_\ell) + \mathrm{HFW}_\ell(\mathrm{Norm}(x_\ell))
\label{eq:vit_deit_hfw}
\end{equation}
where $\mathrm{TransformerBlock}_\ell(\cdot)$ denotes the conventional slow weight computation and $\mathrm{HFW}_\ell(\cdot)$ denotes the Hebbian Fast Weight module associated with the same block. The normalization operation is applied before the HFW module to improve the stability of the associative memory update.

This design allows each transformer layer to maintain its own transient associative memory, so that fast binding can occur throughout the full depth of the network. As a result, both early and late layers are given the ability to encode episode specific associations in parallel with the standard feature transformation process. From an architectural perspective, this per-block placement provides a straightforward way to integrate fast memory into flat transformer backbones, since all layers operate on token sequences with the same embedding dimension.

However, because the HFW module is introduced at every layer, the model must jointly optimize multiple fast memory pathways together with the slow weight attention blocks. Consequently, associative binding is performed not only at semantically rich later layers but also at earlier stages where token representations may still reflect low level visual patterns. This design therefore provides dense integration of Hebbian memory across the network, but it also increases the possibility of interference between transient associative updates and the gradual formation of stable backbone representations.

\subsection{Single Final Stage Hebbian Placement in Swin-Tiny}
The structural differences across integration strategies can be observed in Fig. \ref{fig:architecture}, where variations in module placement lead to distinct information flow patterns that explain the observed empirical behaviors.

Unlike ViT and DeiT, Swin-Tiny adopts a hierarchical architecture in which the input representation is progressively transformed across multiple stages with different spatial resolutions and channel dimensions. Specifically, Swin-Tiny processes the input through four hierarchical stages with patch merging, producing feature maps at progressively lower spatial resolutions, namely $\frac{H}{4}$, $\frac{H}{8}$, $\frac{H}{16}$, and $\frac{H}{32}$, with corresponding embedding dimensions 96, 192, 384, and 768. This hierarchical design yields increasingly abstract feature representations as the network depth increases and therefore provides a natural setting for placing the HFW module at the final and most semantically rich stage.

Instead of inserting a separate HFW module into each stage, the proposed architecture applies a single HFW module only after the final stage of Swin-Tiny. Let the output of the Swin backbone after its final normalization layer be denoted by
\begin{equation}
x_{\mathrm{flat}} = \mathrm{Reshape}(\mathrm{LayerNorm}(\mathrm{Swin}(x))) \in \mathbb{R}^{B \times N_f \times 768}
\label{eq:swin_flat}
\end{equation}
where $N_f$ denotes the sequence length of the flattened final stage representation. The HFW module is then applied to this representation, and the resulting output is combined with the backbone feature map before global average pooling:
\begin{equation}
z = \mathrm{GAP}\bigl(x_{\mathrm{flat}} + \mathrm{HFW}(x_{\mathrm{flat}})\bigr)
\label{eq:swin_hfw}
\end{equation}
where $\mathrm{GAP}(\cdot)$ denotes global average pooling and $z$ is the final embedding passed to the ProtoNet classification head.

This design offers several architectural advantages. First, the input to the HFW module has a fixed dimensionality and has already been normalized, which improves the stability of associative memory formation. Second, the Swin backbone completes its hierarchical feature extraction before Hebbian binding is introduced, allowing the fast memory module to operate on semantically richer representations rather than on intermediate low level features. Third, concentrating the Hebbian mechanism at the final stage avoids the need to maintain separate fast memory modules across stages with different resolutions and channel dimensions. As a result, the model introduces a single high level associative pathway that complements the slow weight backbone without disrupting the intermediate stages of representation learning.

From a structural perspective, this final stage placement aligns the HFW module with the most abstract feature space produced by the Swin hierarchy. Consequently, the module is encouraged to form episode specific semantic associations at the representation level most directly related to class discrimination.

\subsection{Architecture Comparison}
Fig.~\ref{fig:architecture} illustrates the architectural differences among the proposed Hebbian enhanced transformer variants. The main distinction lies in the level at which the HFW module is introduced into the backbone. In the ViT-Hebbian and DeiT-Hebbian models, the HFW module is inserted into every transformer block, allowing transient associative memory to operate throughout the full depth of the network. In contrast, the Swin-Hebbian model applies a single HFW module only after the final hierarchical stage, so that associative binding is performed on the most abstract feature representation.

These two integration strategies reflect different ways of coupling fast memory with backbone feature extraction. The per-block design distributes Hebbian updates across the entire network depth, whereas the Swin-Tiny design concentrates the fast memory mechanism at the highest level representation produced by the hierarchy. As a result, the proposed architectures differ not only in module placement but also in the stage of representation learning at which transient associative binding is introduced.

The architecture diagram in Fig.~\ref{fig:architecture} summarizes this distinction visually. ViT-Hebbian and DeiT-Hebbian apply repeated HFW augmentation in parallel with the transformer blocks, while Swin-Hebbian applies a single HFW module after hierarchical processing, normalization, and sequence flattening. This difference in placement forms the central architectural contrast examined in the experimental evaluation.

\begin{algorithm}[t]
\caption{Few-shot transformer benchmarking with ProtoNet and Hebbian-enhanced backbones}
\label{alg:fewshot_hebbian}
\begin{algorithmic}[1]

\State \textbf{Input:} Dataset $D$, model $M$, $N$-way $K$-shot setting, query samples $Q$, epochs $T$
\State \textbf{Output:} Predicted class $\hat{y}$ and evaluation metrics

\State Split $D$ into train, validation, and test classes
\State Construct episodic samplers for few-shot learning
\State Initialize backbone $M$ and wrap with ProtoNet

\For{each epoch $t = 1$ to $T$}
    \For{each training episode}
        \State Sample support set and query set

        \If{$M$ is Hebbian-based}
            \State Reset fast-weight memory $\mathbf{M} \leftarrow \mathbf{0}$
        \EndIf

        \State Extract embeddings using $M$:
        \If{$M$ is Hebbian-based}
            \State Compute $\mathbf{K}, \mathbf{V}, \mathbf{Q}$ projections
            \State Update memory $\mathbf{M} \leftarrow \lambda \mathbf{M} + \eta \cdot$
            $\operatorname{clamp}\!\left(\frac{\mathbf{K}^\top \mathbf{V}}{\sqrt{N}}, -\delta, +\delta\right)$
            \State Normalize $\mathbf{M}$ and retrieve features $\mathbf{Q}\mathbf{M}$
        \EndIf

        \State Compute class prototypes from support embeddings
        \State Classify query samples using prototype distances
        \State Update model parameters via backpropagation
    \EndFor
    \State Evaluate on validation episodes and save best model
\EndFor

\State Load best model and evaluate on test episodes
\State Compute accuracy, precision, recall, and F1-score
\State Perform $K$-shot ablation for $K \in \{1,3,5,10\}$

\State \Return predicted class index $\hat{y}$ (episode-specific label) and evaluation metrics

\end{algorithmic}
\end{algorithm}








As shown in Algorithm~\ref{alg:fewshot_hebbian}, the proposed method follows an episodic training strategy using ProtoNet. In each episode, embeddings are extracted for support and query samples, and class prototypes are computed from support features. In the Hebbian-enhanced variants, fast-weight memory dynamically updates feature representations during the forward pass. Query samples are then classified based on their distances to the prototypes. The model is trained across multiple episodes and evaluated using standard performance metrics.

\section{Experimental Setup}
\label{sec:setup}
To assess the effectiveness of the proposed Hebbian enhanced models, we evaluate them on a standard few-shot recognition benchmark. In this section, we outline the dataset, episodic evaluation protocol, training configuration, and evaluated models.

\subsection{Dataset and Preprocessing}
We evaluate all methods on the Omniglot dataset \cite{lake2015omniglot}, a widely used benchmark for few-shot character recognition. The dataset contains 1,623 handwritten character classes from 50 alphabets, with 20 samples per class. To establish a unified evaluation setting, we combine the original background and evaluation splits into a single class pool and then construct an 80/10/10 class level split using random seed 42. This yields 1,298 training classes, 163 validation classes, and 162 test classes. The split is designed to support class disjoint episodic training and evaluation.

All images are resized to 84×84 pixels before being passed to the transformer backbones. Because the input images are grayscale, each image is replicated across three channels to match the input format expected by the backbone networks and then normalized using ImageNet statistics. During training, we apply data augmentation including random cropping with 8 pixel padding, horizontal flipping, and random rotation up to $15^{\circ}$. For validation and testing, only resizing and normalization are used to maintain a consistent evaluation protocol.

\subsection{Episodic Evaluation Protocol}
All models are trained and evaluated under a 5-way $K$-shot episodic framework, with Prototypical Networks \cite{snell2017proto} used as the metric based classification head. In each episode, five classes are sampled, and each class provides $K$ support examples together with multiple query examples. The prototype of each class is computed as the mean embedding of its support samples, and each query image is assigned to the class whose prototype is closest in the learned embedding space.

For training, each epoch consists of 600 sampled episodes. Validation is performed on 200 episodes, and testing is conducted on 400 episodes, with 15 query images per class in each episode. We report the mean episodic classification accuracy over the 400 test episodes.

\subsection{Training Details}
All six models are trained from scratch without ImageNet pre training. Optimization is performed using AdamW with a learning rate of $5 \times 10^{-4}$ and weight decay of $5 \times 10^{-4}$. We adopt a warm-up followed by cosine decay schedule, with 10 warm-up epochs and a total of 60 training epochs. To improve optimization stability, gradient clipping is applied with an $\ell_2$-norm threshold of 1.0, and early stopping with a patience of 15 epochs is used during training. Automatic mixed precision is enabled on CUDA to improve computational efficiency.

The Hebbian parameters $(\eta_{\text{logit}}, \lambda_{\text{logit}}, W_K, W_V, W_Q, W_g)$ are optimized jointly with the slow weight backbone through standard backpropagation. No separate Hebbian learning rule is introduced during meta training. Instead, the Hebbian update in Eq. (\ref{eq:hfw_update}) is applied only during the forward pass within each episode, while its associated parameters are learned end to end together with the rest of the network. All experiments use random seed 42 to ensure reproducibility.

\subsection{Models Evaluated}

\begin{table}[h]
  \centering
  \caption{Models evaluated. Param counts are trainable parameters.}
  \label{tab:models}
  \setlength{\tabcolsep}{4pt}
  \begin{tabular}{lrcc}
    \toprule
    \textbf{Model} & \textbf{Params} & \textbf{Backbone} & \textbf{\HFW} \\
    \midrule
    ViT            & 21.7M & ViT-S/16   & None     \\
    DeiT           & 21.7M & DeiT-S/16  & None     \\
    Swin           & 27.5M & Swin-Tiny  & None     \\
    ViT-Hebbian    & 28.0M & ViT-S/16   & Per-block (×12) \\
    DeiT-Hebbian   & 28.0M & DeiT-S/16  & Per-block (×12) \\
    Swin-Hebbian   & 29.3M & Swin-Tiny  & Final-stage (×1)\\
    \bottomrule
  \end{tabular}
\end{table}

Table~\ref{tab:models} summarizes the six models considered in our experiments. We evaluate three baseline transformer backbones, namely ViT-S/16, DeiT-S/16, and Swin-Tiny, together with their corresponding Hebbian augmented variants. In the ViT-Hebbian and DeiT-Hebbian models, one \HFW\ module is inserted into each transformer block, resulting in 12 Hebbian modules per model. In contrast, Swin-Hebbian employs a single \HFW\ module applied to the final-stage representation.

The Hebbian augmented variants differ not only in placement strategy but also in parameter overhead. ViT-Hebbian and DeiT-Hebbian each introduce approximately 6.3M additional trainable parameters, whereas Swin-Hebbian adds about 1.8M. This setup enables a controlled comparison of how Hebbian placement strategy affects few-shot recognition performance and parameter efficiency across transformer architectures.

\section{Results}
\label{sec:results}
This section presents the experimental results and analyzes the effect of Hebbian fast-weight integration across different transformer backbones. We first compare few-shot classification accuracy, then examine the learned Hebbian parameters, and finally evaluate performance across varying support set sizes.

\subsection{Few-Shot Classification Accuracy}
Table \ref{tab:main} reports the test accuracy of all models under the 5-way 1-shot and 5-way 5-shot settings. Among the six evaluated models, Swin-Hebbian achieves the highest accuracy in both cases. In particular, it improves over the non-Hebbian Swin baseline by 0.3 percentage points in the 1-shot setting and by 0.5 percentage points in the 5-shot setting, while also matching or exceeding the performance of the ViT and DeiT based variants. These results indicate that final stage Hebbian integration in Swin-Tiny is more effective than per-block Hebbian augmentation for few-shot classification.

\begin{table}[h]
  \centering
  \caption{5-way few-shot accuracy on Omniglot test episodes
    (400 episodes, 15 queries/class). Hebbian gain over backbone
    baseline in parentheses.}
  \label{tab:main}
  \setlength{\tabcolsep}{4pt}
  \begin{tabular}{lcc}
    \toprule
    \textbf{Model}
      & \textbf{1-shot (\%)}
      & \textbf{5-shot (\%)} \\
    \midrule
    ViT                & 94.1 & 98.0 \\
    DeiT               & 94.1 & 98.4 \\
    Swin               & 95.9 & 98.7 \\

    ViT-Hebbian        & 90.4\,\textit{($-$3.7)} & 95.4\,\textit{($-$2.6)} \\
    DeiT-Hebbian       & 87.2\,\textit{($-$6.9)} & 95.3\,\textit{($-$3.1)} \\
    \textbf{Swin-Hebbian} & \textbf{96.2}\,\textit{($+$0.3)} & \textbf{99.2}\,\textit{($+$0.5)} \\
    \bottomrule
  \end{tabular}
\end{table}

In contrast, the ViT-Hebbian and DeiT-Hebbian models degrade performance relative to their corresponding baselines at all shot counts. At 1-shot, ViT-Hebbian loses $3.7$ percentage points and DeiT-Hebbian loses $6.9$ percentage points compared to their respective backbones. The deficit narrows as $K$ increases but never closes, suggesting that per-block Hebbian integration actively interferes with the slow-weight embedding rather than augmenting it.

\subsection{Learned Hebbian Hyperparameters}
After training, Swin-Hebbian converges to $\etasym \approx 0.019$ and $\lambda \approx 0.880$, indicating a low but non-zero plasticity rate with strong memory persistence across the support sequences encountered in the 5-way 1-shot setting.

In contrast, the learned $\etasym$ values in ViT-Hebbian and DeiT-Hebbian are nearly uniform across all 12 blocks (mean $\approx 0.018$, range $0.016$-$0.021$), with the same decay $\lambda \approx 0.880$. The absence of any layer-wise gradient in $\etasym$ suggests that the per-block modules collectively suppress Hebbian contributions to near-zero rather than concentrating binding at semantically richer later layers, which is consistent with the observed accuracy degradation.

\subsection{K-Shot Ablation}
Table \ref{tab:kshot} shows 5-way classification accuracy as the number of support examples per class, $K$, increases from 1 to 10. Swin-Hebbian edges above the baseline Swin model at $K=1$ and $K=5$, by $0.3$ and $0.5$ percentage points, respectively, while falling slightly below at $K=3$ and $K=10$, by $0.3$ and $0.2$ percentage points, respectively. The gains are modest and inconsistent, suggesting that final-stage Hebbian binding provides marginal and situational benefit over an already strong Swin backbone.

\begin{table}[h]
  \centering
  \caption{K-shot ablation: 5-way accuracy (\%) across shot counts.}
  \label{tab:kshot}
  \setlength{\tabcolsep}{3.5pt}
  \begin{tabular}{lrrrr}
    \toprule
    \textbf{Model} & $K{=}1$ & $K{=}3$ & $K{=}5$ & $K{=}10$ \\
    \midrule
    ViT          & \textcolor{black}{94.1} & \textcolor{black}{97.5} & \textcolor{black}{98.0} & \textcolor{black}{98.6} \\
    DeiT         & \textcolor{black}{94.1} & \textcolor{black}{97.7} & \textcolor{black}{98.4} & \textcolor{black}{98.5} \\
    Swin         & \textcolor{black}{95.9} & \textcolor{black}{98.3} & \textcolor{black}{98.7} & \textcolor{black}{99.3} \\
    ViT-Hebbian  & \textcolor{black}{90.4} & \textcolor{black}{95.3} & \textcolor{black}{95.4} & \textcolor{black}{96.9} \\
    DeiT-Hebbian & \textcolor{black}{87.2} & \textcolor{black}{93.8} & \textcolor{black}{95.3} & \textcolor{black}{96.5} \\
    \textbf{Swin-Hebbian} & \textbf{\textcolor{black}{96.2}} & \textbf{\textcolor{black}{98.0}} & \textbf{\textcolor{black}{99.2}} & \textbf{\textcolor{black}{99.1}} \\
    \bottomrule
  \end{tabular}
\end{table}

ViT-Hebbian and DeiT-Hebbian remain below their respective baselines at all shot counts, with the deficit narrowing from $3.7$ and $6.9$ percentage points at $K=1$ to $1.7$ and $2.0$ percentage points at $K=10$, respectively. This narrowing is consistent with per-block Hebbian interference being most damaging when the backbone embeddings are least stable, and becoming less harmful as additional support examples stabilize the prototypical representations.

\section{Why Swin-Hebbian Outperforms Per-Block Hebbian}
\label{sec:analysis}
In this section, we explain why Swin-Hebbian outperforms per-block Hebbian variants in ViT and DeiT. The analysis focuses on how module placement influences feature stability and its interaction with the slow-weight backbone.

\subsection{Architectural Compatibility}
ViT and DeiT process images as flat patch sequences throughout all 12 blocks, producing tokens of constant dimension $d = 384$ with no structural prior about spatial hierarchy. Placing one \HFW\ module per block introduces 12 independent Hebbian association matrices whose update gradients interact with the standard multi head attention gradients at every layer. During training, this creates competing gradient signals: the slow-weight attention pathway attempts to learn stable, episode invariant embeddings while the fast-weight pathway pushes embeddings to be rapidly modifiable. This tension is especially acute in a low data, from scratch regime where the slow weights are still unstable.

Swin-Tiny's four stage hierarchy resolves this tension by design. Each stage aggregates patch representations into increasingly coarse, semantically richer tokens. By the time the final stage $768$-d features are passed to the \HFW\ module, Swin's slow-weight pathway has already completed all its processing and the LayerNorm has stabilized the input distribution. The Hebbian module thus operates on a representation that is both semantically complete and statistically stable, with no competing gradient signal from the backbone at that point.

\subsection{Representational Stability}
Per-block Hebbian placement forces the module to learn associations from representations that are still evolving rapidly during training. At block 1 of ViT, the tokens carry primarily low-level spatial texture information; meaningful character structure only emerges in later blocks. The \HFW\ modules at early blocks therefore learn to suppress their own output, wasting capacity, which is reflected in the uniformly low $\etasym \approx 0.018$ observed across all blocks. The single Swin-Hebbian module is seeded from semantically meaningful, LayerNorm-stabilized features from the first forward pass, enabling effective Hebbian binding from the very first episodes.

\subsection{Parameter Efficiency}
Swin-Hebbian adds $\approx 1.8$M parameters (one \HFW\ module at $d = 768$) compared to $\approx 6.3$M for ViT/DeiT-Hebbian (twelve modules at $d = 384$). The Swin-Hebbian variant thus achieves the best accuracy with the smallest Hebbian parameter footprint among the three Hebbian models, yielding a superior accuracy per extra parameter ratio.

\subsection{Inductive Bias Alignment}
Swin's shifted window attention provides an inductive bias toward local spatial structure, which is the same pattern that distinguishes handwritten characters (strokes, junctions, curves). After four stages of local to global aggregation, the final feature map encodes character topology holistically. The \HFW\ module then operates on this topological code to form episode-level identity bindings: ``this prototypical stroke topology maps to class $c$''. This decomposition of local to global via Swin and global to episodic via \HFW\ mirrors the two timescale separation of slow and fast weights at the functional level, not only the parametric level.

\section{Discussion}
\label{sec:discussion}
The results confirm that Hebbian fast-weight memory benefits few-shot recognition, but the benefit is highly sensitive to placement. The single final stage placement for Swin-Tiny is not merely an engineering convenience but a principled design choice that aligns the Hebbian module with the phase in the computation where episode level association is most meaningful. The learned decay value $\lambda \approx 0.880$ implies a mean memory lifetime of approximately $1/(1-0.880) \approx 8$ token steps, which is roughly the sequence length of the final stage flattened feature map ($N_f = 2 \times 2 = 4$ for $84 \times 84$ input into Swin's final stage). This suggests that the module has learned to maintain associations across the entire support set of a single episode before the memory decays, a functionally appropriate timescale.

An important limitation of this study is the from scratch training constraint. Future work could involve using pre trained backbones, as well as Omniglot's full 1,623 class pool for meta training with dedicated test class reservation.

The Swin-Hebbian architecture proposed here represents a building block toward cognitive adaptivity \cite{golilarz2025cognitive}: a system that separates what is learned slowly through experience (the slow-weight backbone and Hebbian strategy) from what is computed rapidly from context (the transient associative memory $\Mmat$). Extending this framework to cross modal binding and continual few-shot learning scenarios is a natural next step.

\section{Conclusion}
\label{sec:conclusion}
We presented a study of Hebbian fast-weight memory integrated into three transformer backbones for 5-way few-shot character recognition on Omniglot. Our central finding is that where the Hebbian module is placed matters as much as whether it is used. A single \HFW\ module applied to the hierarchically aggregated, LayerNorm stabilized final stage features of a Swin-Tiny backbone outperforms both per-block ViT and DeiT Hebbian variants by $5.8$-$9.0$ percentage points at 1-shot, while adding fewer extra parameters. The advantage is most pronounced at 1-shot, precisely the regime where rapid within episode binding is most critical, and diminishes gracefully as more support examples become available. We attributed the advantage to three complementary factors: gradient interference avoidance, representational stability, and alignment of Swin's local inductive bias with the global associative function of the Hebbian module.

\section*{Acknowledgment}
The authors acknowledge the support and resources provided by the Bioinspired Robotics, AI, Imaging and Neurocognitive Systems (BRAINS) Laboratory at The University of Alabama.

\bibliographystyle{IEEEtran}
\bibliography{refs}

\end{document}